\documentclass[conference]{IEEEtran}
\IEEEoverridecommandlockouts
\usepackage{cite}
\usepackage{amsmath,amssymb,amsfonts}
\usepackage{algorithmic}
\usepackage{graphicx}
\usepackage{textcomp}
\usepackage{xcolor}
\def\BibTeX{{\rm B\kern-.05em{\sc i\kern-.025em b}\kern-.08em
    T\kern-.1667em\lower.7ex\hbox{E}\kern-.125emX}}
\begin{document}

\title{A Lightweight Attention-Based CNN with Saliency-Map Interpretability for Skin Cancer Detection}

\author{
\begin{tabular}{cc}
\begin{minipage}{0.42\textwidth}
\centering
\textbf{Sanjana Islam Kasfia}\\
Dept. of Computer Science and Engineering\\
International Standard University\\
Dhaka, Bangladesh\\
Email: sanjanaislamkasfia2648@gmail.com
\end{minipage}
&
\begin{minipage}{0.42\textwidth}
\centering
\textbf{Md. Ismiel Hossen Abir}\\
Dept. of Public Health and Life Sciences\\
Canadian University of Bangladesh\\
Dhaka, Bangladesh\\
Email: ismielabir286@gmail.com
\end{minipage}
\\[3.5em]
\begin{minipage}{0.42\textwidth}
\centering
\textbf{AKM Bahalul Haque}\\
Faculty of Science \& Engineering\\
Abo Akademi University\\
Turku, Finland\\
Email: akmbahalul.haque@abo.fi
\end{minipage}
&
\begin{minipage}{0.42\textwidth}
\centering
\textbf{Abir Mahmud Shahariar}\\
Dept. of Computer Science and Engineering\\
International Standard University\\
Dhaka, Bangladesh\\
Email: abirmahmudshahariar@gmail.com
\end{minipage}
\end{tabular}
}

\maketitle

\begin{abstract}
Skin cancer is among the most prevalent and life-threatening forms of cancer worldwide. Early and accurate diagnosis significantly improves patient survival rates. However, manual dermoscopic assessment remains challenging due to visual similarity. This study proposes an Explainable AI-driven framework for automated binary classification of skin lesions using a lightweight attention-based convolutional neural network (CNN-Attention). The model integrates channel attention to focus on informative regions of dermoscopic images while maintaining low computational complexity. A publicly available dataset comprising 2,637 training and 660 test images was used, with balanced class weights. The proposed model achieved 87\% accuracy with 1.95 million parameters and 7.45 MB storage, matching the best baseline (MobileNet) while using approximately half the parameters and storage. ROC-AUC and PR-AUC scores reached 0.9475 and 0.9430, respectively. Additionally, saliency maps were incorporated as an explainable AI technique to visualize regions influencing model decisions. Comparative analysis with VGG16, VGG19, ResNet101, and MobileNet demonstrates the proposed model’s effectiveness in balancing classification results, computational efficiency, and interpretability.

\end{abstract}


\section{Introduction}
Skin cancer is one of the most widespread cancer types globally \cite{Halawani2025-lu}. It refers to various malignant tumors that develop from abnormal skin cell growth \cite{Shetty2022-ec}. The uncontrolled growth of DNA structures within damaged skin cells leads to this condition \cite{Bozkurt2022-xx}. The primary types are basal cell carcinoma, squamous cell carcinoma, and melanoma \cite{Musthafa2024-ay}. Melanoma is the most dangerous form \cite{Bozkurt2022-xx}. Most cases result from exposure to ultraviolet radiation from sunlight \cite{Afza2022-wt}. It spreads quickly and causes high mortality rates \cite{Shah2023-gi}. Early identification of malignant skin lesions is critical \cite{Shah2023-gi}. Doctors must promptly distinguish harmful lesions from harmless ones \cite{Bozkurt2022-xx}. This early recognition is essential for successful treatment. It directly improves patient survival and recovery prospects. Dermoscopic imaging serves as a valuable non-invasive tool \cite{Yue2023-ik}. It provides magnified and illuminated views of lesion structures \cite{Kumar2024-zg}. These structures are often invisible to the bare eyes \cite{Halawani2025-lu}. However, diagnosis remains challenging. Different lesion types appear visually similar \cite{Halawani2025-lu}. Lesions vary in shape, color, texture, and borders \cite{Shah2023-gi}. This variation makes reliable diagnosis difficult. \\
Traditional assessment methods face significant limitations. This has increased focus on computer-aided diagnostic systems for automated skin lesion analysis \cite{Rezaee2024-lr}. These systems perform automated skin lesion analysis. Standard diagnostic methods involve visual examination, dermoscopy, biopsy, and clinical assessment procedures \cite{Bozkurt2022-xx}. These methods present difficulties. These methods require significant time and expert personnel \cite{Asif2024-dm}. Earlier computer-aided systems relied on manual feature engineering \cite{Kumar2024-rn}. These included color, texture, structure, and morphological characteristics \cite{Sulthana2024-mr}. Traditional machine-learning classifiers evaluated these features \cite{Sulthana2024-mr}. However, significant limitations existed. Results depended on the quality and relevance of feature selection. Dermoscopic images from actual patients posed many challenges \cite{Amiri2024-mo}. These challenges included inadequate contrast, hair presence, reflections, shadows, and assorted artifacts \cite{Amiri2024-mo}. As deep learning has advanced, convolutional neural networks (CNNs) have become powerful tools for analyzing skin lesions \cite{Sulthana2024-mr}. These networks automatically learn hierarchical features from image data without manual descriptor engineering \cite{Shetty2022-ec}. Transfer learning with CNNs has proven beneficial for skin lesion classification \cite{Rezaee2024-lr}. Pretrained architectures can effectively classify skin lesions \cite{Musthafa2024-ay}. This is especially valuable when medical image datasets are small.\\
Even with all this progress, there are still difficulties in achieving reliable and effective automated skin lesion classification. Skin lesion datasets often suffer from class imbalance \cite{Bozkurt2022-xx}. Different lesion types share major similarities \cite{Yue2023-ik}. Lesions in the same category show significant variation \cite{Yue2023-ik}. As a result, Classification becomes inherently difficult. Additionally, image quality and acquisition conditions can introduce artifacts and variations \cite{Kumar2024-zg}. These interfere with feature extraction and prediction \cite{Kumar2024-zg}. Traditional deep CNN models use many parameters and demand high computational resources. Balancing accuracy with model size and efficiency becomes difficult. Recent studies have examined attention mechanisms and feature enhancement techniques. These help deep networks prioritize informative regions. They suppress less relevant information. Some researchers have looked into lightweight designs to reduce computational complexity without sacrificing important discriminative properties. These findings advocate the need for models that accurately classify skin lesions while focusing on salient features and reducing computational demands. \\

To address these issues, this study presents an attention-based CNN framework for the automated classification of skin lesions. We also used transfer learning models for comparative analysis. To improve model transparency, we integrated Explainable Artificial Intelligence (XAI) with saliency maps that display which image regions influenced the model’s classification decisions. Additionally, we designed a lightweight CNN-Attention architecture that reduces computational demands while preserving adequate classification performance. We conducted a comparative analysis between our proposed model and the pretrained architecture. The contributions of this study are listed below:\\
i. We developed an attention-based convolutional neural network framework that applies transfer learning for benign and malignant skin lesion categorization.\\
ii. We incorporated an attention mechanism to direct focus toward meaningful regions in dermoscopic images and created a resource-efficient architecture to lower computational requirements.\\
iii. We applied explainable AI, specifically saliency maps, which visualize the regions that influenced model predictions.\\
iv. We did a comparative analysis of the proposed model with the pretrained architecture and measured their performance using standard evaluation metrics.\\

This research is structured across multiple sections, such as the literature review in section 2, methodology in section 3. Section 4 displays the experimental results. Section 5 includes discussion and comparative analysis. Section 6 concludes and then references.

\begin{figure}[!htb]
\centering
\includegraphics[width=0.9\linewidth]{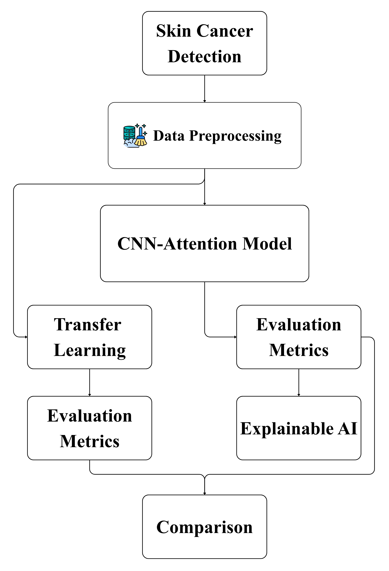}
\caption{Workflow Diagram}
\label{fig:workflow}
\end{figure}

\section{Literature Review}
Recent studies have shown prominent use of several techniques for skin lesion identification. Researchers have used different feature extraction techniques. They have explored various learning architectures. This literature review section summarizes current research approaches.

\subsection{Traditional and Feature-Based Methods}
Salomi et al. developed skin cancer detection system that combines convolutional neural networks, ABCD rule-based feature extraction, and K-means clustering to automatically classify lesions \cite{Salomi2024-wz}. Additionally, they deployed an Android application to ensure timely and widely available skin cancer detection services \cite{Salomi2024-wz}. Hatem et al. constructed a skin lesion classification system using MATLAB and the K-nearest neighbor (KNN) algorithm to classify lesions as normal or malignant \cite{Hatem2022-cf}. His solution confirmed that KNN is a practical approach for accurate and efficient skin lesion classification \cite{Hatem2022-cf}.

\subsection{Deep Learning and Transfer Learning}
Shetty et al. used ML models alongside a custom CNN on the HAM10000 dataset \cite{Shetty2022-ec}. Their custom CNN achieved 95.18\% accuracy \cite{Shetty2022-ec}. The authors propose further advancement through the application of CNN fine-tuning, the use of pretrained models, and the deployment of real-time segmentation and classification methods \cite{Shetty2022-ec}. Bozkurt et al. utilized a pretrained Inception-ResNet-v2 model with affine data augmentation \cite{Bozkurt2022-xx}. Their work addressed skin lesion classification \cite{Bozkurt2022-xx}. Results showed improvement from 83.59\% to 95.09\% accuracy on the augmented dataset \cite{Bozkurt2022-xx}. Their study primarily focused on increasing dataset size and improving classification accuracy \cite{Bozkurt2022-xx}. A hybrid LSTM-CNN framework was developed by Abohashish et al. for binary classification \cite{Abohashish2025-oy}. The model processed image patches to recognize spatial structures and extract local features from the HAM10000 dataset \cite{Abohashish2025-oy}. Shah et al. analyzed the use of ANN and CNN models for automating skin cancer detection \cite{Shah2023-gi}. The review confirmed that both approaches were capable of identifying skin lesions early and with operational capability \cite{Shah2023-gi}. Afza et al. introduced a hierarchical framework that integrates image fusion, superpixel-based lesion segmentation, ResNet-50 model, and feature optimization for skin cancer classification \cite{Afza2022-wt}.

\subsection{Advanced and Efficient Deep Learning Approaches}
Asif et al. developed SKINC-NET, an efficient deep learning model for classifying seven skin lesion types using the HAM10000 dataset \cite{Asif2024-dm}. The model achieved 98.54\% accuracy while requiring less computational power and maintaining fewer trainable parameters than several competing transfer-learning models \cite{Asif2024-dm}. Ali et al. presented an FCEDN-SpaSA-based segmentation method with an adaptive CNN component \cite{Ali2024-uk}. This system distinguishes normal lesions from malignant skin lesions \cite{Ali2024-uk}. The approach reduces energy usage, storage requirements, memory access operations, and training time \cite{Ali2024-uk}. Halawani et al. introduced a hybrid EViT-DenseNet169 model \cite{Halawani2025-lu}. It merges global features obtained from Vision Transformer attention mechanisms with local features extracted from DenseNet169 \cite{Halawani2025-lu}. The model classifies seven categories of skin lesions on the ISIC 2018 dataset \cite{Halawani2025-lu}. Kumar et al. introduced an enhanced Falcon Finch deep convolutional neural network that uses optimization-based parameter adjustment for identifying skin cancer \cite{Kumar2024-zg}. Their study focused on improving accuracy and reducing computational cost \cite{Kumar2024-zg}. Reka et al. investigated quantum ML approaches for skin disease classification using the HAM10000 dataset \cite{Reka2024-ev}. The research incorporated a quantum convolutional neural network and a Quantum Support Vector Classifier, which were then compared against conventional pretrained models \cite{Reka2024-ev}. Kumar et al. developed SCC-Net, a deep CNN model intended for nine-class skin cancer classification on the ISIC-2019 dataset \cite{Kumar2024-rn}. The model incorporates small kernels and dropout regularization to improve computational efficiency and reduce overfitting \cite{Kumar2024-rn}.

\section{Methodology}
This section describes the overall method of CNN with Attention model for skin cancer classification. Figure \ref{fig:workflow} reflects the overall workflow of this study. 

\subsection{Data Analysis and Preparation}

\begin{figure}[!ht]
\centering
\includegraphics[width=0.9\linewidth]{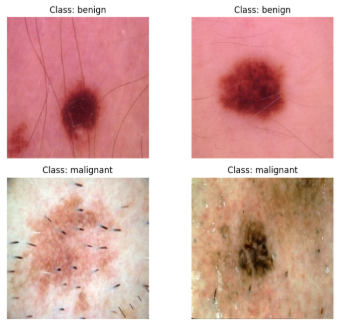}
\caption{Sample Images}
\label{fig:sample_images}
\end{figure}

In this study, we used a public dataset for binary classification of skin lesions \cite{fanconi_skin_cancer}. It consists of dermoscopic images where the train data contains 2,637 images, and the test set contains 660 images. The train data comprises 1,440 benign cases and 1,197 images for malignant cases. Here, the dataset exhibits class imbalance, which was addressed during training through class weighting. All images are RGB and were resized to 224 × 224 pixels. Figure \ref{fig:sample_images} shows sample images from both classes.

The train set of data was divided into two part, such as train and validation subsets. For this task, we used stratified split method where 85\% for training and 15\% for validation. To solve the imbalance issue, we used the class-weight method that were computed using the scikit-learn function. The class weights were approximately 0.915 for the benign class and 1.102 for the malignant class.

\subsection{Proposed CNN-Attention Model}

\begin{figure*}[!htb]
\centering
\includegraphics[width=0.9\linewidth]{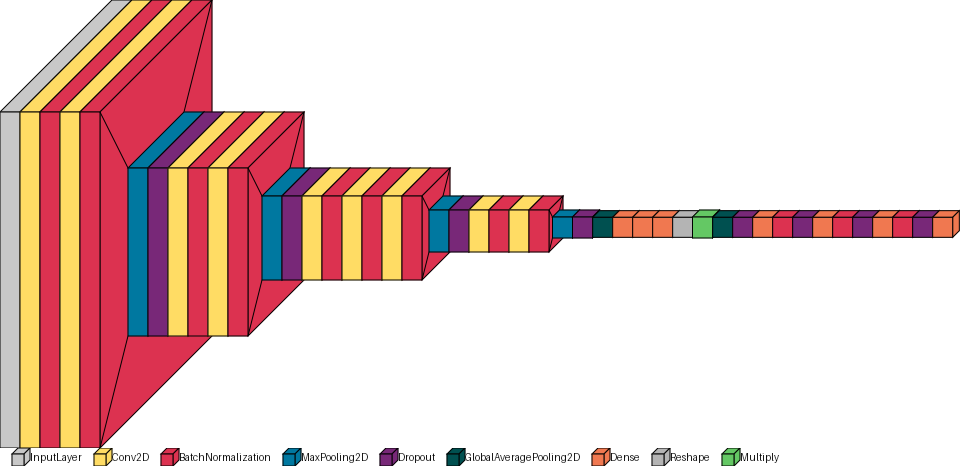}
\caption{Proposed CNN-Attention Model Architecture}
\label{fig:model}
\end{figure*}

We build a custom lightweight CNN-Attention model. The architecture is organized into four convolutional blocks, attention, and then fully connected layers. Figure \ref{fig:model} presents the proposed model. 

The first block contains two 3x3 conv2d layers with 32 filters each. Also, included batchnormalization and for activation purpose we used ReLU. Then, the max-pooling layer contains a 2x2 kernel size to reduce the dimensions, and dropout of 0.2 is applied for regularization. After that, in the second block, we used 64 filters and followed the same pattern, except it increases the dropout rate to 0.25. Similarly, in the third block 128 filters used three convolutional layers, batch normalization, max-pooling, and a dropout rate of 0.3. The fourth block uses 256 filters with two convolutional layers, batch normalization, max-pooling, and dropout of 0.35. All convolutional layers use L2 regularization.

After the convolutional blocks, the channel attention mechanism 
inspired by the Squeeze-and-Excitation (SE) block is used \cite{Hu2017-ue}. The feature map from the last block used global average pooling. Then, we used two fully connected layers with 256 and 512 units, respectively, both using ReLU activation. Then, final dense layer with a sigmoid activation was used. After that, a second global average pooling layer and a classifier head consisting of three dense layers with 512, 256, and 128 units. Each dense layer included batchnormalization and dropout. In the output layer, used a dense layer with softmax activation.

\subsection{Transfer Learning}
To provide a comparative evaluation, four pretrained convolutional neural networks were adapted for the binary classification task: VGG16 \cite{Simonyan2014-wq}, VGG19 \cite{Simonyan2014-wq}, ResNet101 \cite{He2016-fq}, and MobileNet \cite{Howard2017-wy}. In each case, the pretrained backbone was loaded with ImageNet weights, and the feature extractor. Then, a global average pooling layer was used with two dense layers 512 and 256 units. Also, batch normalization and dropout of 0.5 and 0.3 were included, respectively. The final output layer was a dense layer with softmax activation.

\subsection{Explainable AI and Evaluation Metrics}
All models were compiled using the Adam optimizer and initial learning rate of 0.001. We also used sparse categorical cross-entropy for loss function. The proposed CNN-Attention model was trained using 100 epochs, 32 batch size, early stopping (patience 20, restoring best weights), a reduced learning rate on plateau callback (factor 0.2, patience 7), model checkpointing based on validation accuracy, and a custom callback that halted training when validation accuracy reached 99.5\%. The baseline transfer learning models (VGG16, VGG19, ResNet101, and MobileNet) were trained with the same optimizer and loss, but with early stopping patience of 15 and reduced learning rate patience of 5. For training purposes, we used Kaggle notebook environment with GPU. 

For model transparency, an enhanced saliency map was used as an explainable AI technique. Saliency maps were generated by computing the gradient of the predicted class score with respect to the input image to identify influential regions \cite{Simonyan2013-jv}. For evaluation, we used classification report, confusion matrix, ROC AUC, and PR AUC. 

\section{Experimental Results}

Table \ref{tab:classification_report} presents the classification report. For the benign class, precision is 0.91, recall is 0.85, and F1-score is 0.88. For the malignant class, precision is 0.83, recall is 0.89, and F1-score is 0.86. Also, this report shows the accuracy score is 0.87. Both macro and weighted averages are also 0.87.

\begin{table}[!htb]
\centering
\caption{Classification Report for Benign vs Malignant Detection}
\label{tab:classification_report}
\begin{tabular}{lccc}
\hline
\textbf{Class} & \textbf{Precision} & \textbf{Recall} & \textbf{F1-Score} \\
\hline
Benign & 0.91 & 0.85 & 0.88 \\
Malignant & 0.83 & 0.89 & 0.86 \\
\hline
\textbf{Accuracy} & \multicolumn{3}{c}{\textbf{0.87}} \\
\hline
\end{tabular}
\end{table}

\begin{figure}[!htb]
\centering
\includegraphics[width=0.9\linewidth]{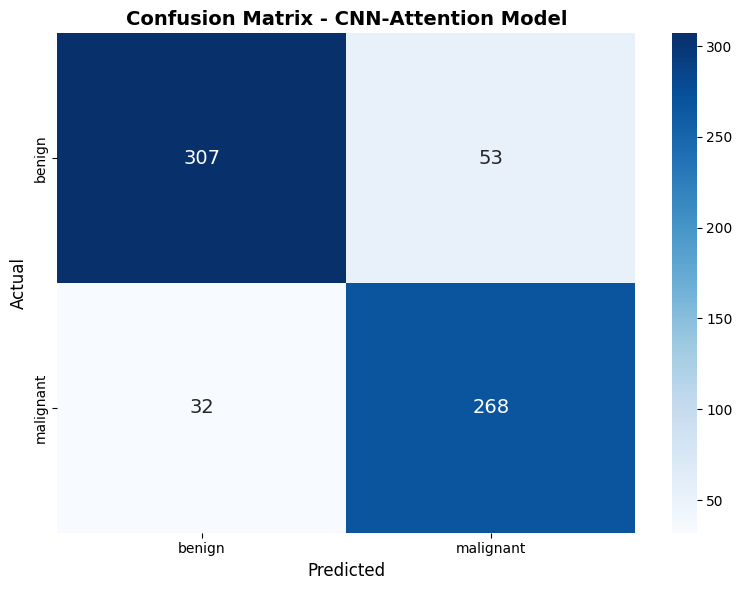}
\caption{Confusion Matrix}
\label{fig:conf}
\end{figure}

Figure \ref{fig:conf} presents the confusion matrix of the CNN-Attention model. It reflects that model correctly predicted 307 benign cases and 268 malignant cases. However, the model shows some misclassifications, with 53 false positives where it predicts benign as malignant. On the other hand, 32 false negative that means malignant identify as benign. Overall, the model shows correct predictions in most cases with an accuracy of 0.87.

\begin{figure}[!ht]
\centering
\includegraphics[width=0.9\linewidth]{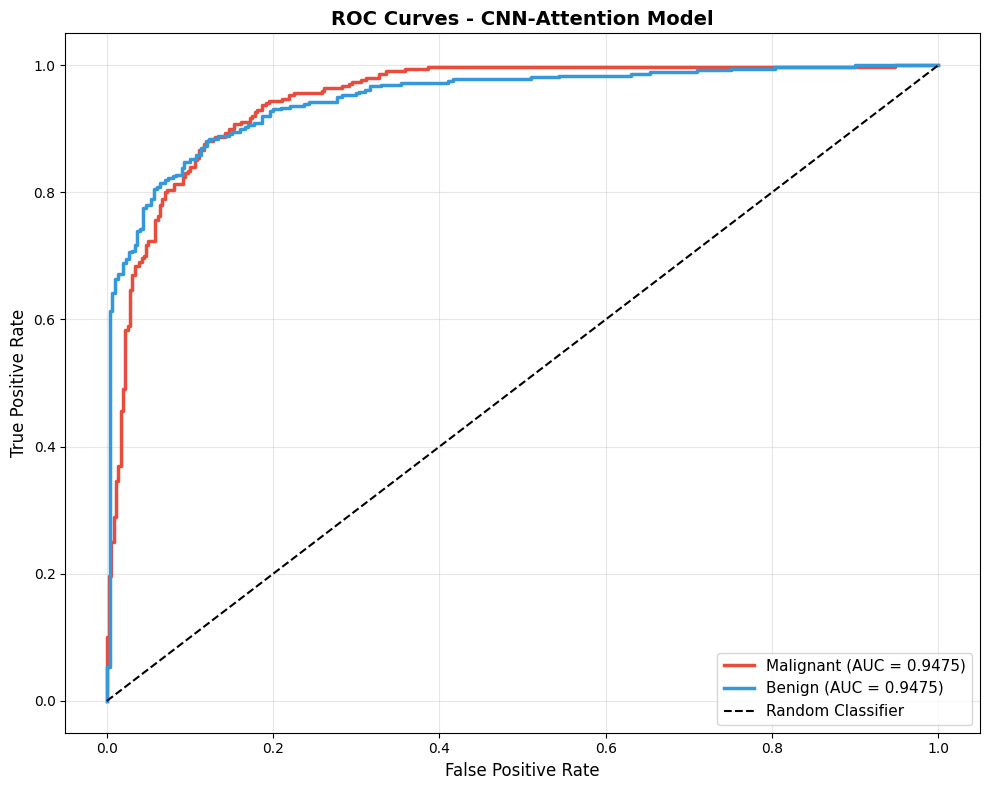}
\caption{ROC}
\label{fig:roc}
\end{figure}

Figure \ref{fig:roc} presents the ROC curves for the CNN-Attention model on both classes. Both the benign and malignant curves have an AUC score of 0.9475. In this case, the model can correctly distinguish benign from malignant cases.

Table \ref{tab:pr_auc} shows the Precision-Recall AUC (PR-AUC) scores for both classes. The benign class has a score of 0.9563, while the malignant class has a score of 0.9296. The average PR-AUC score is 0.9430.

\begin{table}[!htb]
\centering
\caption{Precision-Recall AUC Scores for Benign vs Malignant Detection}
\label{tab:pr_auc}
\begin{tabular}{lc}
\hline
\textbf{Class} & \textbf{PR-AUC Score} \\
\hline
Benign & 0.9563 \\
Malignant & 0.9296 \\
\hline
\textbf{Average} & \textbf{0.9430} \\
\hline
\end{tabular}
\end{table}

\begin{figure}[!htb]
\centering
\includegraphics[width=0.9\linewidth]{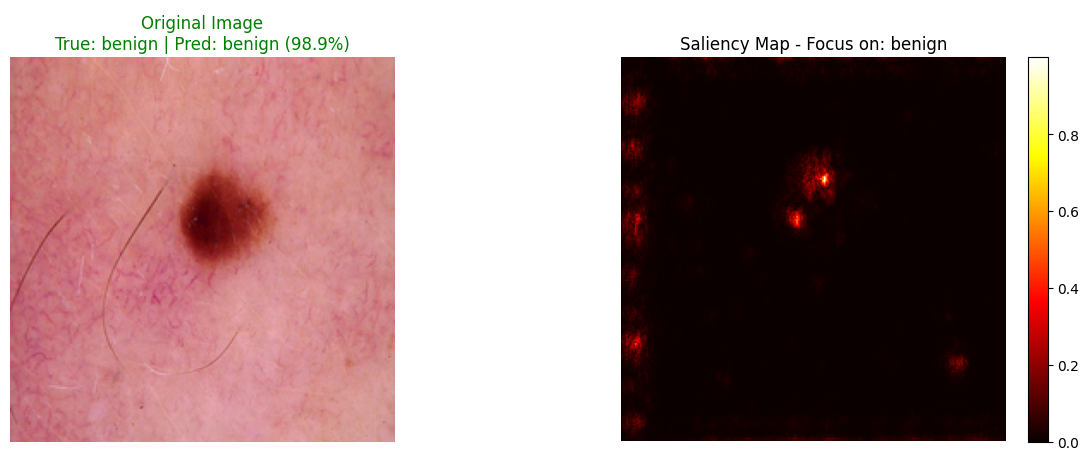}
\caption{Saliency Map for Benign Class}
\label{fig:sam1}
\end{figure}

\begin{figure}[!htb]
\centering
\includegraphics[width=0.9\linewidth]{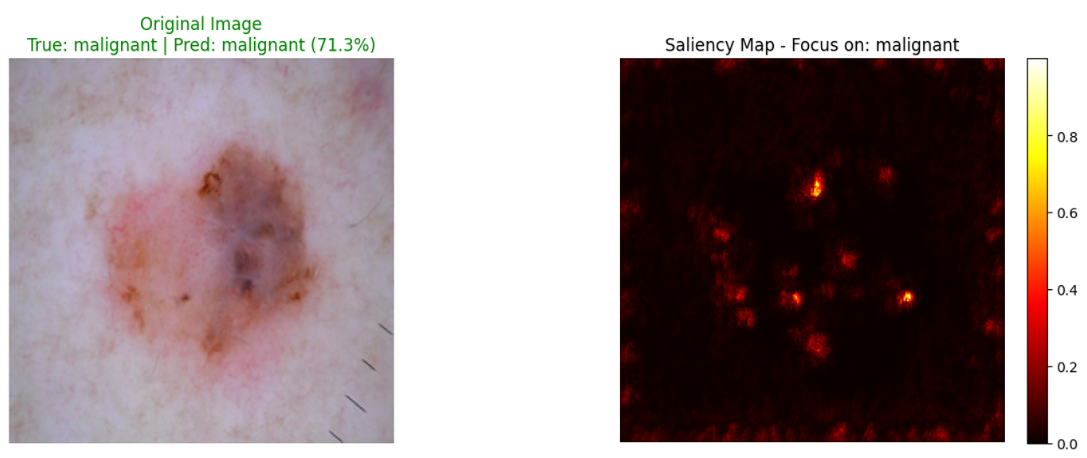}
\caption{Saliency Map for Malignant Class}
\label{fig:sam2}
\end{figure}

Figures \ref{fig:sam1} and \ref{fig:sam2} show the saliency maps of both benign and malignant classes. The left image is truly benign in Figure \ref{fig:sam1} and malignant in Figure \ref{fig:sam2}. 

The right image shows the saliency map. Both images show that the model correctly predicts them as benign or malignant, with confidence scores of 98.9\% and 71.3\%, respectively. The bright yellow and red spots in the saliency maps show the exact regions the model focused on to make a decision for a specific class. In both cases, the model focuses on the actual lesion area rather than the surrounding skin, confirming that its decisions are based on the correct features.

\section{Discussion and Comparative Analysis}

This section presents discussion and comparative analysis of the study. According to Table \ref{tab:model_comparison}, VGG16 and VGG19 both achieved 84\% accuracy, whereas ResNet101 reached 83\% accuracy. MobileNet achieved superior results with 87\% accuracy, the highest among the baseline models. The proposed CNN-Attention model matched this performance level. It also achieved 87\% accuracy. With respect to model complexity, VGG16 and VGG19 contained 15.11 million and 20.42 million parameters, respectively. ResNet101 possessed the largest parameter count at 43.84 million. MobileNet operated with minimal overhead. It required only 3.89 million parameters and consumed 14.84 MB of storage space. The proposed CNN-Attention model involved only 1.95 million parameters and 7.45 MB of storage. The CNN-Attention model provided competitive detection performance. It showed substantially lower parameter usage and storage requirements. The reduced complexity improves its applicability for efficient skin cancer detection.

\begin{table}[!htb]
\centering
\caption{Comparison of Model Performance and Complexity}
\label{tab:model_comparison}
\begin{tabular}{lccc}
\hline
\textbf{Model} & \textbf{Accuracy} & \textbf{Parameters (M)} & \textbf{Size (MB)} \\
\hline
VGG16 & 0.84 & 15.11 & 57.64 \\
VGG19 & 0.84 & 20.42 & 77.90 \\
ResNet101 & 0.83 & 43.84 & 167.25 \\
MobileNet & 0.87 & 3.89 & 14.84 \\
\textbf{CNN-Attention} & \textbf{0.87} & \textbf{1.95} & \textbf{7.45} \\
\hline
\end{tabular}
\end{table}

\section{Conclusion}
In this study, we built a lightweight custom model to detect skin cancer. We classified skin images as benign and malignant. A lightweight CNN-Attention model was developed. It combines transfer learning and an attention mechanism. The mechanism focuses on important regions of skin lesions. The proposed model achieved 87\% accuracy. Precision values were 0.91 for benign and 0.83 for malignant. The average PR-AUC score was 0.9430. The model used only 1.95 million parameters. It required 7.45 MB of storage. This made it smaller than baseline models. It achieved the same accuracy as MobileNet. We integrated Explainable AI using saliency maps. These maps show image regions that influenced classification decisions. Some previous studies reported higher accuracy. However, they did not include Explainable AI. This study merges three important aspects. These are classification performance, lower model complexity, and prediction interpretability. The framework integrates all three. Future studies will use a custom dataset and more advanced techniques. This will improve classification performance. It will further support practical skin cancer detection.

\section*{ACKNOWLEDGEMENT}
During the preparation of this work, the authors used Grammarly Premium, Gemini, and ChatGPT 5.0 to improve sentence structure, fix grammatical mistakes, enhance writing, and maintain coherence throughout the manuscript. After using these, the authors reviewed and edited the content as needed.

\bibliography{references}
\bibliographystyle{ieeetr}

\end{document}